\documentclass{article}

\usepackage{microtype}
\usepackage{graphicx}
\usepackage{subfigure}
\usepackage{booktabs} 

\usepackage{hyperref}

\usepackage[accepted]{icml2024}

\usepackage{amsmath}
\usepackage{amssymb}
\usepackage{mathtools}
\usepackage{amsthm}

\usepackage{amsmath,amsfonts,bm}

\def\eqref#1{equation~\ref{#1}}

\def\1{\bm{1}}

\DeclareMathAlphabet{\mathsfit}{\encodingdefault}{\sfdefault}{m}{sl}
\SetMathAlphabet{\mathsfit}{bold}{\encodingdefault}{\sfdefault}{bx}{n}

\usepackage[capitalize,noabbrev]{cleveref}

\theoremstyle{plain}

\theoremstyle{definition}

\theoremstyle{remark}

\usepackage[textsize=tiny]{todonotes}

\newcommand{\e}[1]{{\small $#1$}}
\usepackage{wrapfig}
\usepackage{multirow,makecell}
\usepackage{tabularx}
\usepackage{enumitem}
\usepackage{pifont}
\usepackage{bbding}
\usepackage{wasysym}
\usepackage{bbm}
\usepackage{listings}
\usepackage{color, colortbl}
\definecolor{Gray}{gray}{0.93}
\definecolor{Orange}{rgb}{1,0.5,0}
\definecolor{DGray}{gray}{0.83}
\definecolor{LightCyan}{rgb}{0.88,1,1}
\usepackage[most, breakable]{tcolorbox}
\usepackage{listings}
\lstdefinestyle{text}{
    basicstyle=\fontsize{8}{9}\ttfamily,
    showstringspaces=false,
    breaklines=true,
    breakatwhitespace=false,
    breakindent=0pt,
    keepspaces=true,
    showspaces=false,  
    escapeinside={(*@}{@*)},
}

\icmltitlerunning{Decomposing Uncertainty for Large Language Models through Input Clarification Ensembling}

\begin{document}

\twocolumn[
\icmltitle{Decomposing Uncertainty for Large Language Models \\through Input Clarification Ensembling}



\icmlsetsymbol{equal}{*}

\begin{icmlauthorlist}
\icmlauthor{Bairu Hou}{ucsb}
\icmlauthor{Yujian Liu}{ucsb}
\icmlauthor{Kaizhi Qian}{ibm}
\icmlauthor{Jacob Andreas}{mit}
\icmlauthor{Shiyu Chang}{ucsb}
\icmlauthor{Yang Zhang}{ibm}
\end{icmlauthorlist}

\icmlaffiliation{ucsb}{UC Santa Barbara}
\icmlaffiliation{ibm}{MIT-IBM Watson AI Lab, IBM Research}
\icmlaffiliation{mit}{MIT CSAIL}

\icmlcorrespondingauthor{Bairu Hou}{bairu@ucsb.edu}

\icmlkeywords{Machine Learning, ICML}

\vskip 0.3in
]



\printAffiliationsAndNotice{}  

\renewcommand{\thefootnote}{\fnsymbol{footnote}} 

\begin{abstract}
Uncertainty decomposition refers to the task of decomposing the total uncertainty of a predictive model into aleatoric (data) uncertainty, resulting from inherent randomness in the data-generating process, and epistemic (model) uncertainty, resulting from missing information in the model's training data. In large language models (LLMs) specifically, identifying sources of uncertainty is an important step toward improving reliability, trustworthiness, and interpretability, but remains an important open research question. 
In this paper, we introduce an uncertainty decomposition framework for LLMs, called input clarification ensembling, which can be applied to any pre-trained LLM.
Our approach generates a set of clarifications for the input, feeds them into an LLM, and ensembles the corresponding predictions. We show that,
when aleatoric uncertainty arises from ambiguity or under-specification in LLM inputs,
this approach makes it possible to factor an (un-clarified) LLM's predictions into separate aleatoric and epistemic terms, using a decomposition similar to the one employed by Bayesian neural networks.
Empirical evaluations demonstrate that input clarification ensembling provides accurate and reliable uncertainty quantification on several language processing tasks.
Code and data are available at \url{https://github.com/UCSB-NLP-Chang/llm_uncertainty}.
\end{abstract}

\renewcommand{\thefootnote}{\arabic{footnote}} 

\section{Introduction}
With the widespread application of large language models (LLMs), it is becoming crucial to ensure predictions from LLMs are trustworthy. One critical dimension of trustworthiness is the ability to indicate when generated text is reliable and correct, which may be formalized as the problem of \textit{uncertainty quantification} (UQ). Uncertainty quantification aims to measure the confidence level of neural networks in their predictions~\citep{gal2016uncertainty, bhatt2021uncertainty, hullermeier2021aleatoric}. A higher uncertainty implies the output of LLMs should be clarified, manually evaluated, or rejected. 

Quantifying LLMs' \emph{total uncertainty} has been the focus of increasing research attention. Existing work observes that LLMs are relatively well-calibrated, especially when predictions are obtained by ensembling multiple reasoning chains~\citep{wang2022self, huang2022large, si2023prompting} or prompts~\citep{jiang2023calibrating}, or when LLMs are prompted to directly output their confidence levels~\citep{kadavath2022language, lin2022teaching, tian2023just}. Many other methods have been proposed to quantify the uncertainty of LLMs~\citep{lin2022teaching,xiao2022uncertainty, kuhn2022semantic, lin2023generating, duan2023shifting, huang2023look, park2023pac, ren2023robots}. Accurate quantification of the uncertainty can be used for various applications, such as out-of-distribution detection and misclassified data detection.

However, \emph{measuring} uncertainty is just the first step towards \emph{understanding} uncertainty in LLM predictions. 
For many applications, it is necessary to distinguish between
different types of uncertainty and decompose the source into these types, a problem we refer to as \emph{uncertainty decomposition}. As discussed more formally below, it is always possible to decompose a predictive model's uncertainty into two components: \emph{aleatoric (data) uncertainty} and \emph{epistemic (model) uncertainty}. Epistemic uncertainty arises when correct outputs are predictable in principle, but models lack the knowledge required for prediction. For example, the question \emph{What is 2+3?} requires the knowledge of algebraic operations. Without such knowledge, the uncertainty will be high.
On the other hand, aleatoric uncertainty arises from ambiguity or inherent randomness in the data-generating process itself: in language processing applications, it may result from ambiguous questions~\citep{min2020ambigqa, guo2021abg, kuhn2023clam} and unclear task instructions~\cite{tamkin2022task}.
In particular, an important source of aleatoric uncertainty is the \emph{input ambiguity}. For example, the answer to the input question \emph{Who is the president of this country} will have a high aleatoric uncertainty because it is ambiguous what country and time the question intends to query about.

This paper aims to obtain a finer-grained uncertainty measurement by determining how much of the total uncertainty can be attributed to aleatoric uncertainty due to input ambiguity.
Aleatoric uncertainty due to input ambiguity is irreducible no matter how well a model learns. For example, to answer the question \emph{Who is the president of this country?}, without any context, the uncertainty would be high regardless of how well the LLM learns, because the question itself is ambiguous. Uncertainty decomposition provides important insights for users to improve the performance of LLM. If epistemic uncertainty is high, users could supply the model with adequate knowledge through model adaptation, in-context learning, \emph{etc.}; if the aleatoric uncertainty is high, then users should modify the query to make it more concrete.

Despite existing work aimed at quantifying total uncertainty in LLMs, decomposing this uncertainty for LLMs remains understudied. Existing methods for uncertainty decomposition in other models cannot be directly applied, due to the black-box nature of LLMs and the prohibitive cost of inference.
For example, Bayesian Neural Networks (BNNs)~\citep{neal2012bayesian, blundell2015weight, graves2011practical, louizos2016structured, hernandez2015probabilistic, hasenclever2017distributed, li2015stochastic} specify a prior distribution over the model parameters and approximate the posterior distribution given the training data. \textsc{Deep Ensembles}~\citep{lakshminarayanan2017simple, fort2019deep} decompose the uncertainty by training different variants of models, \emph{e.g.}, with different random seeds, to with the proper scoring rules (\emph{e.g.}, negative log-likelihood loss for the classification task) in the target task and then ensembling them.

Despite their effectiveness, Bayesian Neural Networks require substantial modifications to the training procedure, while \textsc{Deep Ensembles} necessitate training multiple variants of LLMs. Both approaches are generally impractical or prohibitively expensive. Given these challenges, we aim to address the following question: \textit{How can we effectively quantify and decompose uncertainty in LLMs?}

\begin{figure}[t]
    \centering
    \vspace{1mm}
    \includegraphics[width=.98\linewidth]{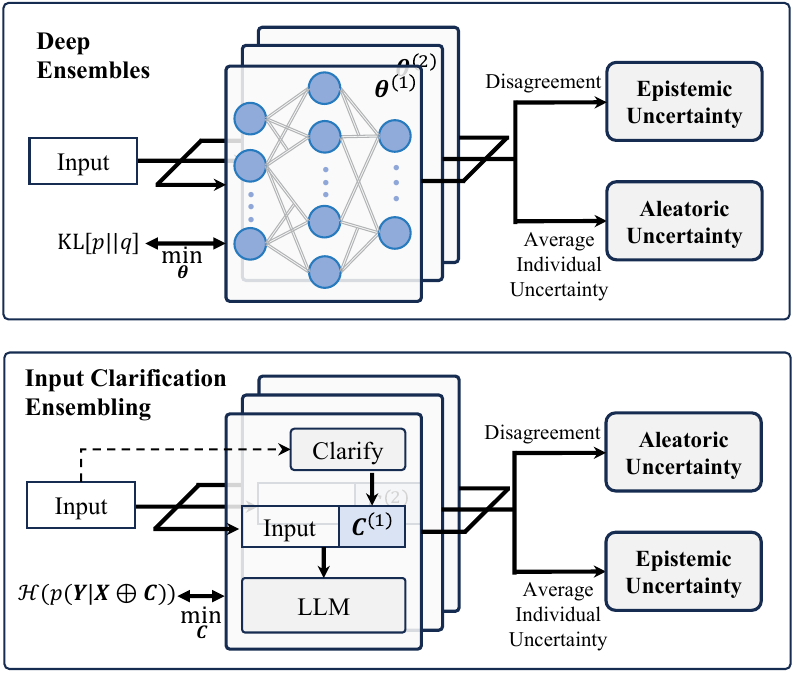}
    \vspace{-1mm}
    \caption{The uncertainty quantification frameworks of \textsc{Deep Ensembles} (upper) and input clarification ensembling (lower).}
    \vspace{-5mm}
    \label{fig:overview}
\end{figure}

In this paper, we propose a framework for uncertainty decomposition that we call \emph{input clarification ensembling}. Our approach shares many intuitions and structural similarities with BNN-based approaches, but  avoid the need to modify LLM parameters or inference procedures. Our approach is motivated by the observation that, although it is very challenging to modify LLM's parameters, it is relatively easy to manipulate the input to LLMs. Inspired by this, rather than ensembling different model variants that minimize the epistemic uncertainty, we introduce a set of \emph{input clarifications} which can minimize the aleatoric uncertainty. We then ensemble an LLM's predictions under different clarifications.  Figure~\ref{fig:overview} shows the general pipeline. 
For example, for the question \emph{Who is the president of this country?}, a possible clarification is \emph{`This country' refers to the US.} By ruling out the aleatoric uncertainty by clarification, we can ascribe the remaining uncertainty of each individual prediction to epistemic uncertainty. Furthermore, by measuring the disagreement of the model predictions under different clarifications, we can gauge the aleatoric uncertainty. Our experiments verify that the proposed method provide accurate uncertainty quantification results on both total uncertainty and its decomposition.

\section{Related Work}
\paragraph{Uncertainty quantification.} Uncertainty quantification for machine learning models has been widely studied to quantify the reliability of model predictions~\citep{gal2016uncertainty, gal2016dropout, malinin2018predictive, ovadia2019can, malinin2020uncertainty, lin2022teaching, kuhn2022semantic, lin2023generating, shen2023post}. 
Uncertainty in model predictions can have numerous causes. Given the total uncertainty in model predictions, one can further decompose it into epistemic uncertainty (due to lack of knowledge in the model) and aleatoric uncertainty (due to the inherent randomness and noise in data). 

Depending on how the uncertainty is obtained, existing uncertainty quantification methods can be categorized into \textit{intrinsic} and \textit{extrinsic} methods. Intrinsic methods adopt machine learning models to provide an inherent uncertainty estimate, such as Bayesian approaches and ensemble-based approaches~\citep{malinin2018predictive}. Bayesian approaches~\citep{blundell2015weight, gal2016dropout, teye2018bayesian, mobiny2021dropconnect,lakshminarayanan2017simple, malinin2020uncertainty, he2020bayesian}  and ensemble-based approaches~\cite{lakshminarayanan2017simple, fort2019deep} can quantify both aleatoric and epistemic uncertainty.
In comparison, extrinsic methods quantify the uncertainty in a post-hoc manner using auxiliary models~\citep{kristiadi2021learnable, lahlou2022deup}. 
Our method belongs to the intrinsic family of methods and is directly motivated by Bayesian neural network approaches.

\paragraph{Uncertainty Quantification and Model Calibration for LLMs}
With the wide application of LLMs, how to accurately quantify the predictive uncertainty has also drawn attention~\citep{xiao2022uncertainty,lin2022teaching, mielke-etal-2022-reducing, zhou2023navigating, huang2023look, duan2023shifting, chen2023quantifying,ott2018analyzing,malinin2020uncertain_auto}. 
Semantic Uncertainty~\citep{kuhn2022semantic} clusters string-valued LLM outputs by synonymy for better uncertainty quantification. 
\citet{lin2023generating} explores uncertainty quantification within the challenging black-box context, where the token generation probability is inaccessible.
In this pursuit, \textsc{BSDetector}~\cite{chen2023quantifying} combines two strategies to estimate the model's predictive uncertainty. The first approach involves sampling multiple answers from the LLM and assessing their consistency, while the second directly queries the LLM for its confidence in the generated answer.
Although there have been some explorations in this direction, existing methods can only estimate the total uncertainty. In comparison, we propose a more principled framework that can both quantify the total uncertainty and decompose it into aleatoric uncertainty and epistemic uncertainty, leading to a more fine-grained understanding of LLMs.

Another line of research is model calibration for LLMs. Model calibration is the process of ensuring that the predicted probabilities or confidence scores from a machine learning model align with the true probabilities or likelihoods of events occurring (\emph{i.e.}, the prediction is correct). Well-calibrated model predictions help improve the reliability of uncertainty quantification.
Using existing model calibration methods~\citep{hendrycks2016baseline,guo2017calibration, ovadia2019can, riquelme2018deep, desai-durrett-2020-calibration}, prior work~\citep{huang2022large, jiang2023calibrating, jiang-etal-2021-know, ye2022explanations} has shown that LLMs are relatively well-calibrated on factual QA and complex reasoning tasks when properly prompted. Specifically, ~\citet{kadavath2022language, tian2023just} estimate the prediction confidence of LLMs by prompting LLMs to output their confidence of their answers. For complex reasoning tasks, LLMs may output both the reasoning chains and the final answer. To estimate the confidence score, previous approaches~\cite {huang2022large} sample multiple outputs for the input question and use the answer frequency to indicate the confidence. Researchers further ensemble multiple prompts for better calibration performance~\citep{jiang2023calibrating}. Our uncertainty quantification is based on the well-calibrated predictions of LLMs, which lead to a more precise and accurate quantification result.

\paragraph{Modeling Ambiguity with language models} Ambiguity is a longstanding issue in the NLP domain, extensively explored in tasks such as syntactic and semantic parsing~\citep{koller2008regular}, open-domain question-answering~\citep{min2020ambigqa, cole2023selectively}, conversational question-answering~\citep{guo2021abg} and natural language inference~\citep{liu2023we}. Prior work, such as \texttt{AmbigQA}~\citep{min2020ambigqa} and \texttt{AmbigEnt}~\citep{liu2023we}, have identified the widespread ambiguities and established benchmarks with ambiguous inputs. 
These studies have demonstrated that existing language models lack the capability to effectively recognize and manage ambiguities. Our work models the ambiguity from the perspective of uncertainty quantification, where a high aleatoric uncertainty can indicate the potential existence of input ambiguity. By decomposing the aleatoric uncertainty, we show that it is possible to enhance the ambiguity detection performance of existing LLMs.

\section{Methodology}
\subsection{Notations and Problem Formulation}

Denote by \e{\bm X} and \e{\bm Y} the input and output target of a given task and \e{\bm \theta} as the parameters of an LLM. Denote by \e{p(\bm Y | \bm X)} and \e{q(\bm Y | \bm X, \bm \theta)} the ground-truth and predicted distribution of \e{\bm Y} given \e{\bm X}.

We first introduce three uncertainty concepts. First, the \emph{\textbf{total uncertainty}} is defined as the entropy of the predicted distribution, \emph{i.e.}, \e{\mathcal{U}_{total} = \mathcal{H}(q(\bm Y | \bm X; \bm \theta))}.
If the overall uncertainty is high, then it means the LLM has low confidence in its output. The total uncertainty can be further decomposed into two different types of uncertainties. 

The first type of uncertainty is referred to as the \emph{\textbf{epistemic uncertainty}}, which characterizes how well the LLM approaches the ground truth distribution, and thus learns the knowledge therein. For example, to answer \emph{`What is 2+3?'}, if the LLM were able to learn the true knowledge of the algebraic operation, it would be able to answer with certainty; otherwise, the uncertainty would be high.

The second type of uncertainty is referred to as the \emph{\textbf{aleatoric uncertainty}}, which characterizes the fundamental uncertainty residing in the ground-truth distribution, and is irreducible no matter how well the LLM learns. For example, to answer \emph{`Who is the president of this country?'}, even if the LLM were well acquainted with politics, it still could not answer it confidently, because this question is inherently ambiguous. The data aleatoric is often quantified by the entropy in the ground-truth distribution, \emph{i.e.}, \e{\mathcal{H}(p(\bm Y | \bm X))}.

The goal of this paper is to estimate both the epistemic and aleatoric uncertainties in LLMs.

\subsection{Background: Bayesian Neural Networks and \textsc{Deep Ensembles}}

The possible solutions to our task is to apply the canonical Bayesian Neural Network (BNN) approach~\cite{blundell2015weight, graves2011practical} or \textsc{Deep Ensembles}~\citep{lakshminarayanan2017simple}, which are standard approaches to uncertainty decomposition.
Instead of having one set of parameters, BNNs model the parameter distribution of a neural network. With the Bayesian formalism, the posterior distribution can be approximated give the training data via techniques such as variational inference~\cite{blundell2015weight, graves2011practical}. Due to the prohibitive
computational cost of BNNs, non-Bayesian methods such as \textsc{Deep Ensembles} are proposed with better scalability. \textsc{Deep Ensembles} maintain \e{K} models, each parameterized as \e{\bm \theta^{(k)}}. Each of the \e{k} models seeks to minimize the training loss, usually the cross entropy loss for classification tasks, which is equivalent to solving the following optimization problem
\begin{equation}
    \small
    \min_{\bm \theta} \mathrm{KL}(p(\bm Y | \bm X) \Vert q(\bm Y | \bm X, \bm \theta))).
    \label{eq:bnn_obj}
\end{equation}

In \textsc{Deep Ensembles}, different models have slightly different initialization values and thus the optimized values, \e{\{\bm \theta^{(k)}\}}, are different.
Denote the resulting distribution of the model parameters \e{\theta} as \e{p(\bm \theta | \mathcal{D})} (either approximated by BNNs or \textsc{Deep Ensembles}) where $\mathcal{D}$ is the training dataset.
Then the ensembled distribution of BNN can be represented as \e{q(\bm Y | \bm X) = \mathbb{E}_{q(\bm \theta | \mathcal{D})}[q(\bm Y | \bm X, \bm \theta)]}. Then we can decompose the predictive uncertainty as

\begin{equation}
    \small
    \mathcal{H}(q(\bm Y | \bm X)) = \underbrace{\mathcal{I}(\bm Y; \bm \theta | \bm X)}_{\text{\ding{172}}} + \underbrace{\mathbb{E}_{q(\bm \theta | \mathcal{D})}\mathcal{H}(q(\bm Y | \bm X, \bm \theta))}_{\text{\ding{173}}},
    \label{eq:bnn_decompose}
\end{equation}
where \e{\mathcal{I}} denotes the mutual information under the \e{q} distribution. \ding{172} measures the disagreement among the different models; \ding{173} measures the average uncertainty of each individual model. 
The above equation can be straightforwardly derived from the definition of conditional mutual information.
Under certain assumptions, \ding{172} and \ding{173} can approximate the epistemic and aleatoric uncertainties, respectively \cite{gal2016uncertainty}.
An illustration of the \textsc{Deep Ensembles} framework is shown in the upper panel of Figure~\ref{fig:overview}.

Here is an intuitive explanation of why this is the case. According to Eq.~\ref{eq:bnn_obj}, the goal of each model is to approach the ground-truth distribution, and thus can be viewed as the process of reducing the epistemic uncertainty. Therefore, if the optimization is successful, all the models will learn the true distribution, \emph{i.e.}, \e{q(\bm Y | \bm X, \bm \theta^{(k)}) = p(\bm Y | \bm X), \forall k}, which, by definition, results in zero epistemic uncertainty. Meanwhile, \ding{172} will also be zero because all the models produce the same prediction. Thus \ding{172} equals epistemic uncertainty in this case. \ding{173} would also equal the aleatoric uncertainty because the predicted distribution is equal to the true distribution.

On the other hand, if the models fail to learn the true distribution, in which case the epistemic uncertainty will be large, \ding{172} will also be large since different models have different hyperparameter settings and will be stuck in very different local optima.

\subsection{Do BNN and \textsc{Deep Ensembles} work for LLMs?}
\label{sec: bnn_analyese_for_llm}

Our goal of decomposing uncertainty for LLMs would be easily achieved if these methods were readily applicable to LLMs. Unfortunately, this is not the case. For BNNs, we need to significantly modify the training method of LLMs. For \textsc{Deep Ensembles}, the learning process in Eq.~\ref{eq:bnn_obj} is also very challenging for LLMs. Specifically, there are two types of methods for adapting LLMs to a particular task, supervised fine-tuning and prompting/in-context learning. Directly fine-tuning the model according to Eq.~\ref{eq:bnn_obj} is usually infeasible due to the limited access to model parameters and its huge requirement for computation. Even if it is feasible, it would be very time-consuming because it requires fine-tuning multiple LLMs.

\begin{figure}[h]
    \centering
    \hspace{-5mm}
    \includegraphics[width=.85\linewidth]{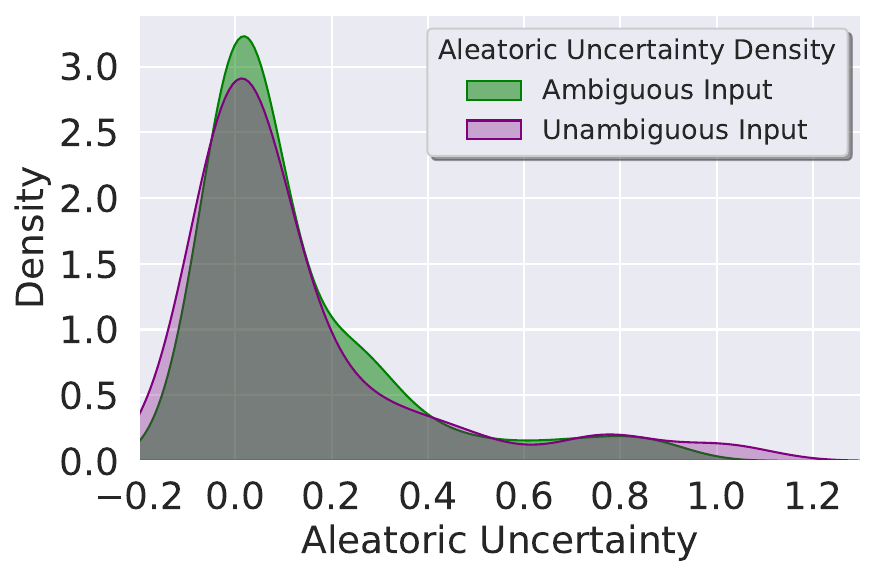}
    \vspace{-3mm}
    \caption{Aleatoric uncertainty distribution on the \texttt{AmbigQA}~\citep{min2020ambigqa} dataset using the \textsc{Deep Ensembles} method. We use kernel density estimation to smooth the frequency distribution histogram. \textsc{Deep Ensembles} is achieved by ensembling different in-context examples.}
    \label{fig: bnn_ours_cmp}
    \vspace{-3mm}
\end{figure}

On the other hand, the in-context learning method, though feasible, does not fit into the \textsc{Deep Ensembles} framework because it does not directly aim to optimize Eq.~\ref{eq:bnn_obj}, so the decomposition will be very inaccurate. To demonstrate this, we perform a simple experiment on the \texttt{AmbigQA}~\citep{min2020ambigqa} dataset, which contains both ambiguous questions with multiple answers and unambiguous questions. We use the BNN method to decompose the uncertainty of ChatGPT, where the different individual model is derived by providing different in-context examples. If the decomposition method is accurate, we would expect to see that the aleatoric uncertainty for the ambiguous questions is significantly larger than that of the unambiguous ones. However, as shown in Figure~\ref{fig: bnn_ours_cmp}, the gap between the uncertainties of the two groups of questions is very small. More experiment details can be found in Section~\ref{sec: exp}.

While the BNN and \textsc{Deep Ensemble} framework do not work for LLMs, it inspires us to design an alternative framework that is almost completely symmetrical to the BNN approach, as discussed in the next subsection.

\subsection{Input Clarification Ensembling}

Although modifying or adapting LLMs is challenging, it is relatively straightforward to modify the input to LLMs. By analogy to the way BNNs and \textsc{Deep Ensembles} ensemble different \emph{models} that minimize \emph{epistemic uncertainty} (Eq.~\ref{eq:bnn_obj}), can we design a framework that ensembles different \emph{inputs} to minimize \emph{aleatoric uncertainty}? 

This is the motivation behind our framework, which consists of the following two steps.

\noindent \textbf{Step 1: Input Clarification.} Given an input \e{\bm X}, we first generate a set of texts, \e{\bm C^{(k)}}, called \emph{clarifications}. Each clarification \e{\bm C^{(k)}} seeks to minimize the ambiguity in \e{\bm X} (and thus the aleatoric uncertainty) when appended to \e{\bm X}. Formally, we denote one clarification result as \e{\bm X \oplus \bm C^{k}}, 
where \e{\oplus} denotes concatenation. In the aforementioned example, \emph{`Who is the president of this country?'}, possible clarifications include \emph{`This country refers to the US.'} and many other countries. Since there can be many valid clarifications for the input, \e{\{\bm C^{(k)}\}} is a set.

\noindent \textbf{Step 2: Ensemble.} We denote the distribution of the aforementioned input clarifications as \e{q(\bm C | \bm X)} given a particular input \e{\bm X}. 
Then, we define the predictive model $q(\bm Y | \bm X)$
as an ensemble of predictions conditional on each clarified input, \emph{i.e.}, \e{q(\bm Y | \bm X) = \mathbb{E}_{q(\bm C | \bm X)}[q(\bm Y | \bm X \oplus \bm C, \bm \theta)]}. (Model parameters \e{\bm \theta} are kept constant, and thus will be omitted for brevity below.)

We then propose to decompose the uncertainty of the ensembled model as
\begin{equation}
    \small
    \mathcal{H}(q(\bm Y | \bm X)) = \underbrace{\mathcal{I}(\bm Y; \bm C | \bm X)}_{\text{\ding{172}}'} + \underbrace{\mathbb{E}_{q(\bm C | \bm X)}\mathcal{H}(q(\bm Y | \bm X \oplus \bm C))}_{\text{\ding{173}}'}.
    \label{eq:our_decompose}
\end{equation}
We claim that \ding{172}$'$, which computes the mutual information between the model output distribution and the clarifications, can approximate the aleatoric uncertainty caused by input ambiguity. In contrast, \ding{173}$'$ is the average entropy of the output distribution given different clarifications, representing the model’s uncertainty across clarified versions of the input.
Assuming that input ambiguity is the sole source of aleatoric uncertainty, we may consider it as an estimate of the epistemic uncertainty for the LLM given the original input. This interpretation, however, diverges from the traditional definition of epistemic uncertainty, and we mainly focus on decomposing the aleatoric uncertainty (\ding{172}$'$) in this paper.

By comparing the above process against Eqs.~\ref{eq:bnn_obj} and \ref{eq:bnn_decompose}, 
we can notice the symmetry between our framework and \textsc{Deep Ensembles}'s --- \textsc{Deep Ensembles} seeks to pin down epistemic uncertainty whereas ours aleatoric uncertainty; Eq.~\ref{eq:our_decompose} takes almost an identical form to Eq.~\ref{eq:bnn_decompose} but the corresponding uncertainties are swapped. Figure~\ref{fig:overview} also shows such symmetry.

Accordingly, the same explanation of why it works applies here. When the input is already very clear, and hence aleatoric uncertainty is low, the clarifications will be identically empty, so \ding{172}$'$ will approach zero. When the input is very ambiguous, the clarifications will be very different (think about the aforementioned president example), and so would the answers produced with different clarifications. In this case, \ding{172}$'$ will be very high. On the other hand, \ding{173}$'$ measures the average uncertainty on \emph{clarified} input, which rules out most of the aleatoric uncertainty, so the remaining uncertainty can mostly be ascribed to epistemic uncertainty.

\subsection{Input Clarification}
\label{subsec:input_clarification}

Unlike the conventional neural networks, the input to LLMs usually contains multiple components, including instructions, in-context examples, questions \emph{etc}. Therefore, we can separately measure the aleatoric uncertainties caused by different input components by clarifying only the corresponding components. For example, to measure the aleatoric uncertainty resulting from ambiguous instructions, we can clarify only the instruction. 
For the aleatoric uncertainty studied in this work, we will focus on the uncertainty caused by instruction ambiguity and question ambiguity, but the framework is readily generalizable to other input components.

To derive clarifications that approximately minimize the ambiguity in step 1 above,
we introduce a clarification LLM, where we provide an instruction and in-context example to guide the LLM to perform adequate clarification.
Therefore, the above input clarification distribution  \e{q(\bm C | \bm X)} in Equation~\ref{eq:our_decompose} is the output distribution of the clarification LLM.
Note that the clarification LLM can be different from the LLM for prediction. 
In this work, we propose the following design choices for the clarification LLM to to ensure the quality of clarification:
\vspace{-0.05in}
\begin{itemize}[leftmargin=*]
    \item Prompting an LLM with task instructions and in-context examples. We design instructions for the clarification generation task and provide the model (\texttt{gpt-3.5-turbo} and \texttt{gpt-4}) with several in-context examples.
    \item Supervised fine-tuning. We can also fine-tune a open-sourced language model on datasets that contains the ambiguous inputs and their corresponding clarifications. We fine-tune the \texttt{Llama-3-8b-instruct} model on the training set of the \texttt{AmbigQA}~\cite{min2020ambigqa} dataset.
\end{itemize}
\vspace{-0.05in}
Further implementation details are provided in Section~\ref{sec: exp} and Appendix~\ref{sec: fine_tuning}. For both design choices, we show that we can easily adapt the LLMs for clarification generation and quantify the uncertainty using our proposed framework.

\subsection{Improving Performance via Soliciting Clarifications}
\label{subsec:human_interaction}

Our framework not only provides a way of decomposing the uncertainties, but can also enable an interpretable and effective human-LLM interaction experience. Currently, one of the major ways for humans to interact with LLMs is designing appropriate input. However, the input designed by humans may not be clear enough to LLMs, often resulting in undesirable answers given by LLMs. With the proposed input clarification framework, we can design an interaction paradigm that alleviates this problem.

Given an input query, we can first gauge the uncertainties of different input components. If one of the components, say the instruction, contributes to high uncertainty (exceeding a threshold), we can provide feedback to the user that the LLM is not sure about the answer because the instruction is ambiguous, along with several clarification options produced by the clarification LLM for the user to choose from. This would help the user to perform directed improvement of the input query and obtain the desirable answer.

\section{Experiments}
\label{sec: exp}
In this section, we conduct empirical evaluations to demonstrate the validity and effectiveness of the proposed method. Specifically, we aims to answer the following two questions:
\begin{enumerate}
    \item Can the proposed UQ framework quantify \textit{total uncertainty} effectively and correctly?
    \item Can the proposed UQ framework \textit{decompose the uncertainty} effectively and correctly?
\end{enumerate}
To answer the first question, we conduct the mistake detection experiment, which will be introduced in Section~\ref{subsec: mistake}. To answer the second question, we conduct three experiments: ambiguity detection, monotonicity check, and recall of correct answers, which will be presented in Sections~\ref{subsec: ambig_detect}-\ref{subsec: recall}, respectively.

\subsection{Experiment Configurations}
\label{subsec: exp_config}
We use \texttt{gpt-3.5-turbo-0613} as the default LLM for all experiments. We sample 10 model predictions with temperature \texttt{0.5} and use the answer frequency to estimate the output distribution. Since all the datasets we use are open-ended generation tasks, different generated answers could have the exactly same meaning. For example, to answer the question \emph{`When did the world's population reach 7 billion?'}, the LLM may generate several different answers such as \emph{`December 2017'} and \emph{`The world's population reached 7 billion in December 2017'}, which are essentially the same meaning. Regarding these two answers as distinct answers can lead to an overestimation of the entropy of output distribution. Previous work~\citep{kuhn2022semantic, lin2023generating} uses a natural language inference model to cluster different generated sequences with the same semantic meanings into one group for better output distribution estimation. We empirically find that LLMs can achieve better clustering performance. Therefore, we prompt the LLM to cluster output answers into different groups for output distribution estimation on question-answering datasets. More details about this process can be found in Appendix~\ref{app: llm_cluster}.

For all the experiments, we introduce the following baselines:
Semantic Uncertainty~\citep{kuhn2022semantic} (denoted as \textsc{SE}) directly computes the entropy of the output distribution as the estimated (total) uncertainty (named \emph{semantic entropy} in their paper). ~\citet{tian2023just} first queries the LLM for the answer and then queries the LLM again for the confidence of the correctness of the answer. We denote this method as \textsc{Ask4Conf}. We also slightly modify the prompt for the ambiguity detection task to query LLM for the confidence of the ambiguity of the input (denoted as \textsc{Ask4Conf-d}).
The \textsc{Deep Ensembles} method (denoted as \textsc{Ensembles}$^*$ for brevity) is implemented by ensembling the output distributions of multiple different in-context example sets (we use 5 different sets). We add $^*$ here since this method is different from standard \textsc{Deep Ensembles} and does not directly optimize Eq.~\ref{eq:bnn_obj}.
We provide more details of the prompts used in the experiments in Appendix~\ref{subsec:prompt_detail}.

\subsection{Quantifying Total Uncertainty}
\label{subsec: mistake}
Correctly quantifying the total uncertainty is the premise of correctly decomposing the uncertainty. If the estimated total uncertainty is inaccurate, so will the estimated aleatoric and epistemic uncertainty. A reliable total uncertainty measure should have a close correspondence to the model's prediction accuracy. For model predictions whose total uncertainty is high, the chances that the predictions are incorrect should also be high. Therefore, we will evaluate the total uncertainty quantification using the mistake detection task, following the previous work~\citep{kuhn2022semantic, lin2023generating}.

\paragraph{Evaluation Settings}

We evaluate the total uncertainty on the \texttt{Natural Question} (\texttt{NQ}) dataset \citep{kwiatkowski2019natural} and \texttt{GSM8K}~\citep{cobbe2021training}. For each dataset, we randomly sample 200 examples from the validation set for evaluation. The total uncertainty on each example is used to predict whether the model's answer is correct. We report the area under the receiver operator characteristic curve (AUROC) as well as the best F1 score when using the total uncertainty to predict the correctness of the model answer. We use 5-shot in-context examples on the \texttt{NQ} dataset and 2-shot on the \texttt{GSM8K} dataset with chain-of-thoughts. We prompt the LLM to rephrase the input question to generate the clarification set. The detailed prompts are listed in Appendix \ref{subsec:prompt_detail}.

\paragraph{Results} 
The experiment results are shown in Table~\ref{tab: exp_mistake_detection}, which confirms that the total uncertainty measured by the proposed approach is reliable. 
Specifically, we highlight the following observations. First, our method achieves comparable uncertainty quantification performance compared to the baselines, achieving a similar AUROC and F1 score. Second, as the proposed method shares a symmetry form with the \textsc{Deep Ensembles} method, one would expect the total uncertainty quantification of the two should be similar. The above experimental results verify that the quantification results of these two methods are very close. Third, although \textsc{Ask4Conf} performs well on factual QA tasks, it provides a poor uncertainty estimation for the complex reasoning task (\texttt{GSM8K}), while our method can still provide good mistake detection performance.

\newcommand{\orangecolor}[1]{\textcolor{lightorange}{#1}}
\newcommand{\gr}{\rowcolor[gray]{.95}}

\begin{table}[t]
    \centering
    \vspace{-2mm}
    \caption{Uncertainty quantification for mistake detection.  Entropy (\ding{52}) refers to the average total uncertainty of questions with correct answers, while Entropy (\ding{56}) refers to the average total uncertainty of question with wrong answers.}
    \vspace{0.1in}
    \definecolor{rulecolor}{RGB}{0,71,171}
    \definecolor{tableheadcolor}{RGB}{204,229,255}
    \newcommand{\myrowcolour}{\rowcolor{tableheadcolor}}
    \newcommand{\highest}[1]{\textcolor{blue}{\textbf{#1}}}
    \newcommand{\topline}{ %
        \arrayrulecolor{rulecolor}\specialrule{0.1em}{\abovetopsep}{0pt}%
        \arrayrulecolor{rulecolor}\specialrule{\lightrulewidth}{0pt}{0pt}%
        \arrayrulecolor{tableheadcolor}\specialrule{\aboverulesep}{0pt}{0pt}%
        \arrayrulecolor{rulecolor}
        }
    \newcommand{\midtopline}{
        \arrayrulecolor{tableheadcolor}\specialrule{\aboverulesep}{0pt}{0pt}%
        \arrayrulecolor{rulecolor}\specialrule{\lightrulewidth}{0pt}{0pt}%
        \arrayrulecolor{white}\specialrule{\aboverulesep}{0pt}{0pt}%
        \arrayrulecolor{rulecolor}}
        \newcommand{\bottomline}{
        \arrayrulecolor{tableheadcolor}\specialrule{\aboverulesep}{0pt}{0pt}
        \arrayrulecolor{rulecolor}
        \specialrule{\heavyrulewidth}{0pt}{\belowbottomsep}
        \arrayrulecolor{rulecolor}\specialrule{\lightrulewidth}{0pt}{0pt}
        }
    \newcolumntype{?}{!{\vrule width 1.4pt}}
    \resizebox{0.48\textwidth}{!}{
    \begin{tabular}{lcccc}
    \topline
    \textbf{Method} & \textbf{AUROC} & \textbf{F1 Score} & \textbf{Entropy (\ding{52})} & \textbf{Entropy (\ding{56})} \\
    \midtopline \myrowcolour
\multicolumn{5}{c}{\textbf{\texttt{Natural Question}}} \\ \midtopline
        \textsc{Semantic Entropy}  & 63.8 & 77.9  & 0.29 & 0.56 \\
        \textsc{Ask4Conf}  & 70.4 & 83.9  & - & -  \\
        \textsc{Ensembles}$^*$  & 69.7 & 79.7  & 0.46 & 0.88  \\
        \gr \textsc{Ours}  & 72.3 & 80.2 & 0.58 & 1.18  \\
    \midtopline \myrowcolour
\multicolumn{5}{c}{\textbf{\texttt{GSM8K}}} \\ \midtopline
    \textsc{Semantic Entropy}  & 88.2 & 92.4 & 0.32 & 1.46 \\
    \textsc{Ask4Conf}  & 58.1 & 92.3  & - & -  \\
    \textsc{Ensembles}$^*$  & 88.3 & 94.6   & 0.57 & 1.94  \\
        \gr \textsc{Ours} & 89.7 & 94.7 & 0.42 & 1.82  \\
\midtopline
\bottomline
    \end{tabular}
    }
    \label{tab: exp_mistake_detection}
    \vspace{-5mm}
\end{table}

\subsection{Uncertainty Decomposition}
\label{subsec: ambig_detect}
Now we can proceed to evaluate whether the decomposed uncertainty is reliable. As discussed, one of the main causes of aleatoric uncertainty is the ambiguity of the input. Therefore, we will test how well the measured aleatoric uncertainty is predictive of whether an input is ambiguous. In particular, we focus on two input components, the instruction and the question, and separately predict the ambiguity within each component using the respective aleatoric uncertainty (see Section~\ref{subsec:input_clarification}).

\paragraph{Datasets}

For ambiguity detection of the question, we select the \texttt{AmbigQA} dataset~\citep{min2020ambigqa}, which has annotations on the ambiguity of questions. 
The questions in \texttt{AmbigQA} are extracted from the \texttt{NQ} dataset~\citep{kwiatkowski2019natural}.
For ambiguity detection of the instruction, since there is no existing dataset, we create a dataset, \texttt{AmbigInst}, where we generate ambiguous instructions, their disambiguation, and the input-output pairs using ChatGPT. Each instruction is paired with around 15 questions. Since the focus of \texttt{AmbigInst} is to detect ambiguous instructions, we do not introduce ambiguity to the paired questions. More details about \texttt{AmbigInst} can be found in Appendix~\ref{sec: syn_dataset}. We use the full \texttt{AmbigInst} dataset and randomly sample 200 examples from the validation set of \texttt{AmbigQA} for evaluation.

\paragraph{Evaluation Settings}
We use 5-shot in-context examples on the \texttt{AmbigQA} dataset similar to the experiment on the \texttt{NQ} dataset. Since the questions in \texttt{AmbigInst} are relatively easy and straightforward, we directly prompt LLMs in a zero-shot setting.
For ambiguous question detection, we perform clarifications on the input question only. 
We evaluate two clarification LLMs on the \texttt{AmbigQA} dataset, including the \texttt{Llama-3-8B-Instruct} fine-tuned on the training set of \texttt{AmbigQA} and \texttt{gpt-4}. we retrieve the most similar 16 questions as in-context examples when prompting the \texttt{gpt-4} to generate clarifications for a particular input question.
The similarity between two questions is measured by the cosine similarity of their sentence embeddings from \textsc{Sentence-BERT}\footnote{We use the pre-trained sentence transformer model https://huggingface.co/sentence-transformers/all-mpnet-base-v2.}~\citep{reimers2019sentence}. For the \texttt{AmbigInst} dataset, we directly prompt \texttt{gpt-3.5-turbo-0613} to generate instruction clarifications (See Appendix \ref{subsec:prompt_detail} for more details). 
We also include the performance of our method when using ground-truth disambiguation from the two datasets for reference (denoted as \textsc{Ours}$^*$).

The baselines are similar to the methods in the mistake detection task. The main difference is we use the quantified uncertainty to predict whether the input contains ambiguity. The total uncertainty for \textsc{SE} is used for ambiguity prediction in this task. Also, we test both the aleatoric uncertainty and total uncertainty quantified by the \textsc{Deep Ensembles} method, denoted by \textsc{Ensembles}$^*$(aleatoric) and \textsc{Ensembles}$^*$ (total) respectively. For our method, we use the aleatoric uncertainty for ambiguity prediction. \textsc{Deep Ensembles} is not included on the \texttt{AmbigInst} dataset since we do not include in-context examples on that dataset. We also incorporate results with additional methods from previous work~\citep{si2022prompting, cole2023selectively} in Appendix~\ref{subsec: additional_results}.

\begin{table}[t]
    \centering
    \vspace{-2mm}
    \caption{Uncertainty quantification for ambiguity detection.  Avg. AU (\ding{52}) refers to the average aleatoric uncertainty of unambiguous questions, while Avg. AU (\ding{56}) refers to the average aleatoric uncertainty of ambiguous questions.}
    \vspace{0.1in}
    \definecolor{rulecolor}{RGB}{0,71,171}
    \definecolor{tableheadcolor}{RGB}{204,229,255}
    \newcommand{\myrowcolour}{\rowcolor{tableheadcolor}}
    \newcommand{\highest}[1]{\textcolor{blue}{\textbf{#1}}}
    \newcommand{\topline}{ %
        \arrayrulecolor{rulecolor}\specialrule{0.1em}{\abovetopsep}{0pt}%
        \arrayrulecolor{rulecolor}\specialrule{\lightrulewidth}{0pt}{0pt}%
        \arrayrulecolor{tableheadcolor}\specialrule{\aboverulesep}{0pt}{0pt}%
        \arrayrulecolor{rulecolor}
        }
    \newcommand{\midtopline}{
        \arrayrulecolor{tableheadcolor}\specialrule{\aboverulesep}{0pt}{0pt}%
        \arrayrulecolor{rulecolor}\specialrule{\lightrulewidth}{0pt}{0pt}%
        \arrayrulecolor{white}\specialrule{\aboverulesep}{0pt}{0pt}%
        \arrayrulecolor{rulecolor}}
        \newcommand{\bottomline}{
        \arrayrulecolor{tableheadcolor}\specialrule{\aboverulesep}{0pt}{0pt}
        \arrayrulecolor{rulecolor}
        \specialrule{\heavyrulewidth}{0pt}{\belowbottomsep}
        \arrayrulecolor{rulecolor}\specialrule{\lightrulewidth}{0pt}{0pt}
        }
    \newcolumntype{?}{!{\vrule width 1.4pt}}
    \resizebox{0.48\textwidth}{!}{
    \begin{tabular}{lccccc}
    \topline
    \textbf{Method} & \textbf{AUROC} & \textbf{F1 Score} & \textbf{Avg. AU (\ding{52})} & \textbf{Avg. AU (\ding{56})} \\
    \midtopline \myrowcolour
    
    \multicolumn{5}{c}{\textbf{\texttt{AmbigQA}}} \\ \midtopline
    \textsc{Semantic Entropy}  & 54.9 & 46.8 & 0.24 & 0.47 \\
    \textsc{Ask4Conf-d}  & 55.0 & 64.3 & - & -  \\
    \textsc{Ensembles}$^*$ (aleatoric)  & 53.6 & 53.0 & 0.13 & 0.13  \\
    \textsc{Ensembles}$^*$ (total) & 55.4 & 55.0 & 0.50 & 0.41  \\
    \gr \textsc{Ours} (GPT) & 71.7 & 70.1 & 0.28 &  0.67 \\     
    \gr \textsc{Ours} (LLaMA) & 67.1 & 71.8 & 0.55 &  0.91 \\     
    \gr \textsc{Ours}$^*$  & 89.8 & 85.6 & 0.53 & 1.52  \\
    \midtopline \myrowcolour
    \multicolumn{5}{c}{\textbf{\texttt{AmbigInst}}} \\ \midtopline
    \textsc{Semantic Entropy}  & 66.0 & 53.7 & 0.07 & 0.50 \\
    \textsc{Ask4Conf-d}  & 57.9 & 75.4 & - & -  \\
    \gr \textsc{Ours} (GPT) & 81.3 & 77.9 & 0.10 & 0.75  \\
    \gr \textsc{Ours}$^*$  & 96.7 & 92.6 & 0.10 & 1.04  \\
\midtopline
\bottomline
    \end{tabular}
    }
    \label{tab: exp_ambig_detection}
    \vspace{-4mm}
\end{table}

\paragraph{Results} The experiment results are shown in Table~\ref{tab: exp_ambig_detection}. We emphasize two observations. First, our method achieves the best ambiguity detection performance and significantly outperforms the baselines. 
Note that all the baselines, except for \textsc{Ensembles}$^*$ (aleatoric), use the total uncertainty for ambiguity detection, and thus could not disentangle epistemic uncertainty from the aleatoric uncertainty. 
Therefore, these results verify the importance of uncertainty decomposition.
Second, even a small model can also be efficiently adapted for the clarification generation task. When using the fine-tuned LLaMA model, Our method can still outperform baselines significantly. This adaptation process was remarkably efficient, requiring less than 10 minutes on 4$\times$80G H100 GPUs.
Third, the \textsc{Deep Ensembles}$^*$ (aleatoric) method is not effective in the black-box LLM setting. As we have discussed in Section~\ref{sec: bnn_analyese_for_llm}, simply varying the in-context examples cannot accurately estimate the parameter posterior distribution, while the proposed framework is specially designed for the black-box LLMs.

Another observation is that ambiguity detection performance varies across different datasets. On the \texttt{AmbigQA} dataset~\citep{min2020ambigqa}, the ambiguities are more implicit and hard to find by the clarification models, which makes the detection performance relatively low (although still higher than baselines significantly). \citet{min2020ambigqa} also note that the ambiguity in the dataset is ``sometimes subtle'' and ``many (ambiguities) are only apparent after examining one or more Wikipedia pages''.
In comparison, on the \texttt{AmbigInst} dataset where we design ambiguities to be very explicit (see Appendix~\ref{sec: syn_dataset} for more examples), the clarification model can generate effective clarifications for most cases, leading to a good detection performance. Finally, the performance of our method can be further improved when using with the ground-truth disambiguation from the two datasets as the input clarifications, demonstrating that the clarification model is still worth exploring.

\begin{figure}[t]
    \centering
    \vspace{1mm}
    \includegraphics[width=.9\linewidth]{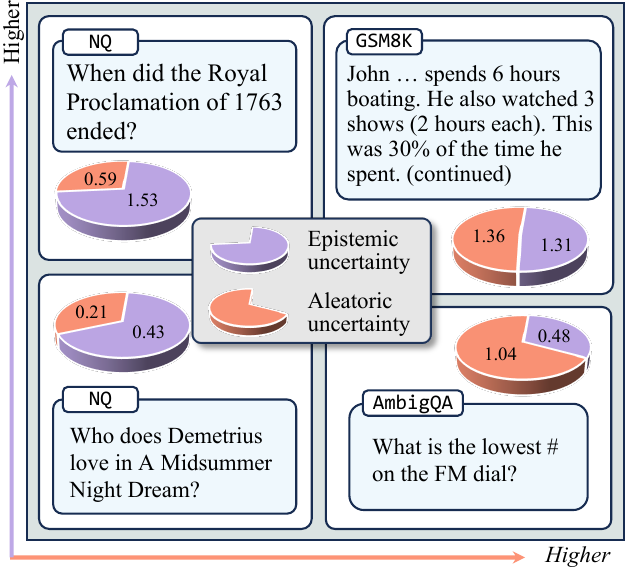}
    \vspace{-2mm}
    \caption{The uncertainty quantification examples using the proposed method. The instances are selected from existing datasets including \texttt{Natural Question (NQ)}~\citep{kwiatkowski2019natural}, \texttt{AmbigQA}~\citep{min2020ambigqa}, and \texttt{GSM8K}~\citep{cobbe2021training}.}
    \vspace{-3mm}
    \label{fig:qualitative_example}
\end{figure}

\paragraph{Qualitative Results}
We further present a visual representation of various uncertainty quantification results in Figure~\ref{fig:qualitative_example}. These examples have been grouped according to the levels of aleatoric and epistemic uncertainty the LLM exhibits. Our uncertainty quantification framework enables a clear understanding of the sources of uncertainty in each example. For instance, consider the question ``\texttt{What is the lowest \# on the FM dial}'' from the \texttt{AmbigQA} dataset. This question lacks specificity regarding the country and region, leading to ambiguity in the input. Our uncertainty quantification illustrates that the predominant source of uncertainty in the model's prediction stems from aleatoric uncertainty in this case. In contrast, despite a clear question, the model struggles to answer a query about the ``Royal Proclamation" (as shown in the upper left example), resulting in a high level of epistemic uncertainty. Interestingly, we have identified a few examples within the \texttt{GSM8K} dataset where the uncertainty in the LLM prediction is attributed to data-related factors. For instance, the upper right example in the figure raises ambiguity about whether the word ``\texttt{this}'' refers solely to watching shows or encompasses both activities (the ground-truth annotation uses the second interpretation).
More details about these examples can be found in Appendix~\ref{app: qualitative}.

\subsection{Monotonicity Check}
\label{subsec: monotonicity}

To further evaluate the reliability of our aleatoric uncertainty measure, particularly the clarification module, we perform a monotonicity check experiment. Ideally, the clarified input should contribute to a much lower aleatoric uncertainty than the original ambiguous input. To test this, we perform two rounds of aleatoric uncertainty measuring. In the first round, we measure the aleatoric uncertainty by clarifying the original input segments (question or instruction). In the second round, we measure the aleatoric uncertainty of the clarified inputs obtained in the first round. Our goal is to check whether the aleatoric uncertainty measured in the second round is much smaller than that in the first round. This experiment is performed on the \texttt{AmbigQA} and \texttt{AmbigInst} datasets. In both rounds, we use the same clarification prompt to generate the clarifications.

\begin{figure}[t]
    \centering
    \hspace{-1mm}
    \subfigure
    {          
        \includegraphics[height=.40\linewidth]{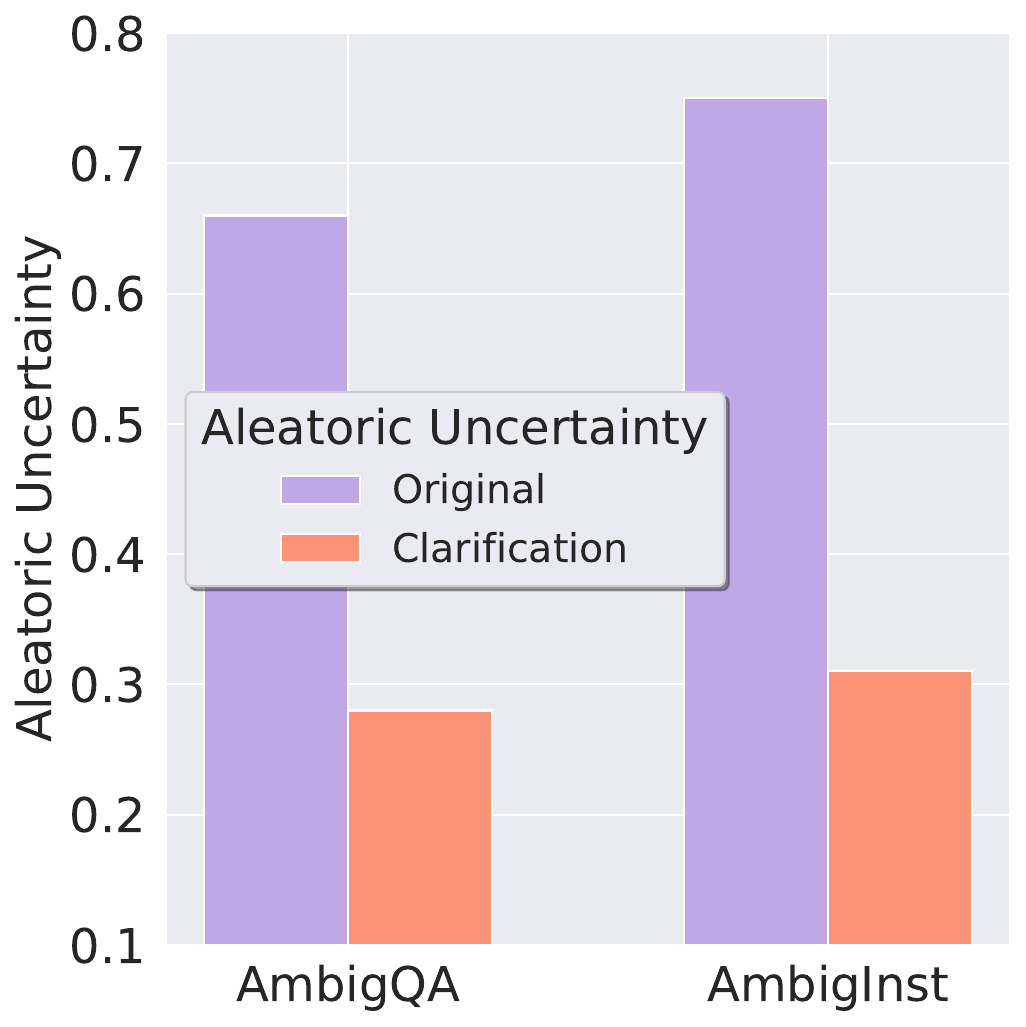}
    }
    \hspace{-1mm}
    \subfigure
    {          
        \includegraphics[height=.40\linewidth]{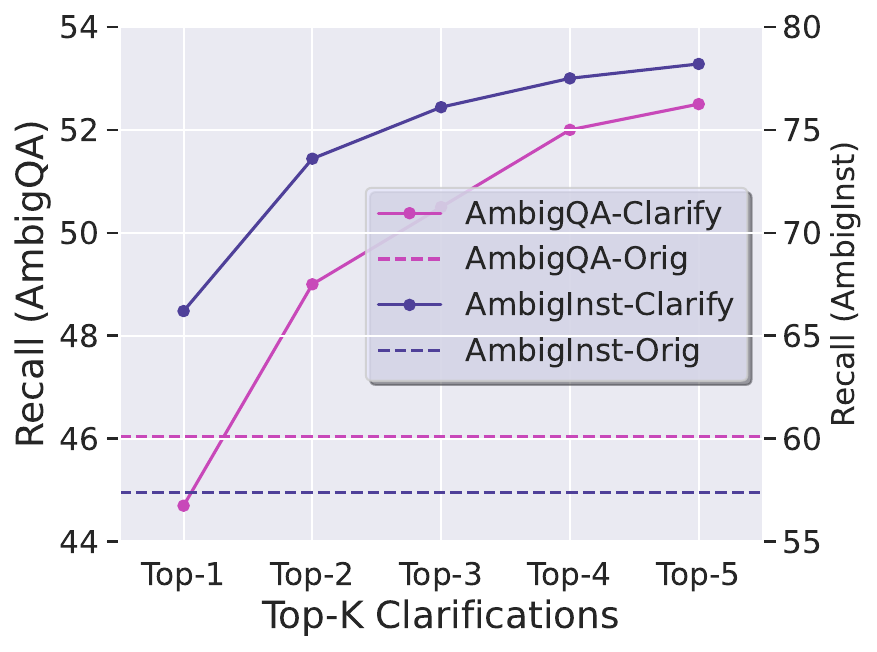}
    }
    \vspace{-2mm}
    \caption{(Left) Average aleatoric uncertainty of the ambiguous inputs and their clarifications. (Right) Performance improvement via Soliciting clarifications. \texttt{AmbigQA-Orig} and \texttt{AmbigInst-Orig} refer to the recall of correct answers when directly answering the original input.  \texttt{AmbigQA-Clarify} and \texttt{AmbigInst-Clarify} refer to the recall of correct answers using different number of input clarifications.}
    \label{fig:recall_improve}
    \vspace{-3mm}
\end{figure}

Figure~\ref{fig:recall_improve}(a) visualizes the change in uncertainty on both datasets. As can be observed, the aleatoric uncertainty drops significantly after the input is clarified, which verifies the effectiveness of the clarification network.

\subsection{Recall of Correct Answers}
\label{subsec: recall}
As discussed in Section~\ref{subsec:human_interaction}, our framework can be used to improve the performance in the presence of ambiguous input by asking users to choose from a set of clarified versions of the input. To make this happen, our methods must be able to cover a good proportion of the possible answers resulting from different clarifications of a given ambiguous input. Also, the number of required clarifications should be smaller, as the users might not want to select the responses from a large set of choices.

To test this, we use the ambiguous questions and instructions from \texttt{AmbigQA} and \texttt{AmbigInst} respectively. For each input, we collect all the possible labeled answers from the ground-truth annotations. Then we select one answer as the target answer that the user is asking for. In our pipeline, the LLM will generate multiple answers given the generated clarifications. Therefore, we inspect how well these generated answers cover the target answer given different numbers of clarifications. We separately compute the recall of the target answer with the different numbers of clarifications. As a baseline, we introduce a vanilla version, where we directly query the LLM with ambiguous input without any clarification.

The results are illustrated in Figure~\ref{fig:recall_improve}(b). We can consistently observe an increase of recall given more clarifications. Similar to the ambiguity detection performance, the recall improvement on the \texttt{AmbigInst} dataset is more significant compared to the \texttt{AmbigQA} dataset, which is due to the subtlety of the \texttt{AmbigQA} dataset as discussed. Nevertheless, the proposed clarification framework is able to significantly improve the answer recall over the vanilla version without the clarification.

\section{Conclusion}
In this paper, we focus on the uncertainty quantification of LLMs and propose a new framework for decomposing the uncertainty. With a symmetric structure of the BNN methods, our framework leverages input clarifications for uncertainty quantification, which is more suitable for black-box LLMs. experimental results affirm that our proposed method not only yields reliable uncertainty quantification but also effectively decompose the total uncertainty into aleatoric and epistemic uncertainty.
In the future, we will further explore how to build a more effective clarification module to boost the effectiveness of our method.

\section*{Acknowledgment}
The work of Bairu Hou, Yujian Liu, and Shiyu Chang was partially supported by National Science Foundation (NSF) Grant IIS-2207052, NSF Grant IIS-2302730, and CAHSI-Google Research Award.

\section*{Impact Statement}
In this paper, our primary objective is to develop an innovative uncertainty quantification framework, which aims to empower users in pinpointing the sources of uncertainty in LLMs accurately.
The principal application scenario for our approach is to improve the trustworthiness and reliability of LLMs. Consequently, the likelihood of unintended usage or potential risks arising from our proposed method is considerably reduced. We also assess both the evaluations and datasets to ensure they are devoid of any harmful content or adverse impacts.

Nonetheless, it's imperative to note that our method relies on the well-calibrated nature of LLMs when applied to factoid QA and mathematical reasoning tasks. A well-calibrated LLM helps improve the reliability of uncertainty quantification. There exists a possibility that LLMs may not exhibit the same level of calibration on particular downstream tasks, and we are committed to continually refining the calibration of LLMs to bolster the reliability and trustworthiness of machine learning systems.

\bibliography{ref}
\bibliographystyle{icml2024}

\newpage
\appendix
\onecolumn

\section{Additional Results and Implementation Details}
\subsection{Additional Results}
\label{subsec: additional_results}
In this section, we include additional methods that estimate the model's confidence on its answer to study whether the confidence score can be used for ambiguity detection. Specifically, ~\citet{si2022prompting}
uses a self-consistency frequency to estimate the LLM confidence (denoted as \textsc{SelfConsistency}), where multiple answers are sampled from the LLM given the original question and the highest answer frequency are used as the confidence score. In addition, ~\citet{cole2023selectively} incorporate two features to estimates the LLM confidence: the answer repetition and the answer diversity. The answer repetition parallels the implementation of ~\citet{si2022prompting}, and the answer diversity is estimated by counting the number of unique answers from multiple sampled answers. We denote the two methods as \textsc{Sample Repetition} and \textsc{Sample Diversity}. We include these 3 confidence scores for ambiguity detection.

\begin{wraptable}{r}{.6\textwidth}
    \centering
    \vspace{-5mm}
    \caption{Additional results for ambiguity detection.  Avg. AU (\ding{52}) refers to the average aleatoric uncertainty of unambiguous questions, while Avg. AU (\ding{56}) refers to the average aleatoric uncertainty of ambiguous questions.}
    \vspace{0.1in}
    \definecolor{rulecolor}{RGB}{0,71,171}
    \definecolor{tableheadcolor}{RGB}{204,229,255}
    \newcommand{\myrowcolour}{\rowcolor{tableheadcolor}}
    \newcommand{\highest}[1]{\textcolor{blue}{\textbf{#1}}}
    \newcommand{\topline}{ %
        \arrayrulecolor{rulecolor}\specialrule{0.1em}{\abovetopsep}{0pt}%
        \arrayrulecolor{rulecolor}\specialrule{\lightrulewidth}{0pt}{0pt}%
        \arrayrulecolor{tableheadcolor}\specialrule{\aboverulesep}{0pt}{0pt}%
        \arrayrulecolor{rulecolor}
        }
    \newcommand{\midtopline}{
        \arrayrulecolor{tableheadcolor}\specialrule{\aboverulesep}{0pt}{0pt}%
        \arrayrulecolor{rulecolor}\specialrule{\lightrulewidth}{0pt}{0pt}%
        \arrayrulecolor{white}\specialrule{\aboverulesep}{0pt}{0pt}%
        \arrayrulecolor{rulecolor}}
        \newcommand{\bottomline}{
        \arrayrulecolor{tableheadcolor}\specialrule{\aboverulesep}{0pt}{0pt}
        \arrayrulecolor{rulecolor}
        \specialrule{\heavyrulewidth}{0pt}{\belowbottomsep}
        \arrayrulecolor{rulecolor}\specialrule{\lightrulewidth}{0pt}{0pt}
        }
    \newcolumntype{?}{!{\vrule width 1.4pt}}
    \resizebox{0.6\textwidth}{!}{
    \begin{tabular}{lccccc}
    \topline
    \textbf{Method} & \textbf{AUROC} & \textbf{F1 Score} & \textbf{Avg. AU (\ding{52})} & \textbf{Avg. AU (\ding{56})} \\
    \midtopline \myrowcolour
    
    \multicolumn{5}{c}{\textbf{\texttt{AmbigQA}}} \\ \midtopline
    \textsc{Semantic Entropy}  & 54.9 & 46.8 & 0.24 & 0.47 \\
    \textsc{Ask4Conf-d}  & 55.0 & 64.3 & - & -  \\
    \textsc{Ensembles}$^*$ (aleatoric)  & 53.6 & 53.0 & 0.13 & 0.13  \\
    \textsc{Ensembles}$^*$ (total) & 55.4 & 55.0 & 0.50 & 0.41  \\
    \textsc{SelfConsistency}  & 56.0 & 62.5 & - & -  \\
    \textsc{Sample Repetition}  & 58.6 & 69.0 & - & -  \\
    \textsc{Sample Diversity}  & 57.6 & 66.7 & - & -  \\
    \gr \textsc{Ours} (GPT) & 71.7 & 70.1 & 0.28 &  0.67 \\     
    \gr \textsc{Ours} (LLaMA) & 67.1 & 71.8 & 0.55 &  0.91 \\     
    \gr \textsc{Ours}$^*$  & 89.8 & 85.6 & 0.53 & 1.52  \\
\midtopline
\bottomline
    \end{tabular}
    }
    \label{tab: additional_results}
    \vspace{-4mm}
\end{wraptable}

We implement these methods as follows. For \textsc{Selfconsistency}, we use the default hyperparameters in the official implementation (temperature = 0.7, sample 10 answers from the model) to compute the confidence. When predict the input ambiguities, we use 1-confidence as the input, since a lower confidence implies either the model's answer is wrong or there are multiple valid answers (i.e., the input is ambiguous).
For \textsc{SampleDiversity}, we use the default hyperparameters in the official implementation (temperature = 0.5, sample 10 answers from the model). We computed the number of unique answers, using this metric to gauge the ambiguity of the input question.
For \textsc{SampleRepetition}, this method parallels \textsc{Selfconsistency}, where answers are generated using greedy decoding and then re-sampled (temperature = 0.5, sample 10 answers) to assess the initial answer's confidence.The experiment results is as below:

We visualize the results in Table~\ref{tab: additional_results}. These results underscore that model confidence alone is insufficient for accurately identifying ambiguous inputs, as evidenced by the AUROC scores of all three methods being under 60. We will also include these findings and citations in the final version of the paper.

\subsection{Supervised Fine-tuning for Clarification Generation}
\label{sec: fine_tuning}
We fine-tuning the \texttt{Llama-3-8B-Instruction} on the full training set of \texttt{AmbigQA} dataset on 4$\times$NVIDIA H100 80GB HBM3 GPU. We organize the training data using the template in Figure~\ref{sft_prompt}. We use \texttt{PyTorch Lightning}, \texttt{DeepSpeed} Stage 1, and \texttt{flash-attention 2} to train the model. We train the model with batch size 16, learning rate 2e-5, and cosine learning rate scheduler for 5 epochs. The loss is only computed on the output tokens. We evaluate the model on the validation set and take the model that achieves lowest validation loss (epoch = 2) for testing.

\begin{figure*}[h]
\begin{tcolorbox}
\begin{lstlisting}[style=text]
<|begin_of_text|><|start_header_id|>user<|end_header_id|>

In this task, you will receive a question that may contain ambiguities. First analyze the following aspects to find if there is any ambiguities according to the real-world facts:
- entities, objects, or events has multiple references or interpretations
- Unclear timestamps
- Unclear locations
- Unclear answer types (e.g., "When" refers to "which year or what date", and "Who" refers to "which person or which team")

If there is any ambiguities, you need to remove ambiguities by adding some clarifications to the question. Each clarification is an additional condition or explanations to the concept in the question that resolve its ambiguity. 
    - You are only allowed to add conditions or explanations, and you cannot change the original intent or semantics of the question.
    - The conditions and explanations must be ground to real-word facts.

If there is no ambiguities, you only need to output the original question as it is.<|eot_id|><|start_header_id|>assistant<|end_header_id|>

Sure. Please provide me with the question so that I can identify whether it is ambiguous and clarify it.<|eot_id|><|start_header_id|>user<|end_header_id|>

Original Question: {original_question}<|eot_id|><|start_header_id|>assistant<|end_header_id|>

Question after adding condition: {ground_truth_clarification}<|eot_id|>
\end{lstlisting}
\end{tcolorbox}
\vspace{-1mm}
\caption{The prompt template for the fine-tuning of \texttt{Llama-3-8B-Instruction}. The \texttt{\{original\_question\}} and \texttt{\{ground\_truth\_clarification\}} are two placeholders that will be filled with the original question (either ambiguous or not) and the ground-truth clarifications.}
\label{sft_prompt}
\end{figure*}


\subsection{Implementation details for baselines}
\label{sec: implementation}
\paragraph{Mistake detection} For the mistake detection task, we strictly follow the experiment settings from \citet{kuhn2022semantic} and \citet{lin2023generating}. For each example, we estimate the output distribution and take the answer with the highest frequency as the final answer. Then we use the method (and the prompt) from \citet{lin2023generating} to determine whether the answer is correct by prompting ChatGPT. Based on the total uncertainty and correctness of the answer, we compute the AUROC and conduct a grid search to find the best threshold for the F1 score, where the correct answers are regarded as positive examples.

For the implementation of \textsc{Ask4Conf}\citep{tian2023just} in the mistake detection task, we use the ``Verb. 2S top-1'' method (and the corresponding prompts) to estimate the confidence of the language model. Rather than asking the LLM to directly generate an answer, we sample multiple answers and take the most frequent one as the answer. After that, we prompt the LLM for the confidence of the most frequent answer. The prompt we use is:

\begin{figure*}[h]
\begin{tcolorbox}
\begin{lstlisting}[style=text]
Answer the following question.
Question: {The testing question}
Answer: {The most frequent answer}

Provide the probability that your answer is correct. Give ONLY the probability, no other words or explanation.

For example:

Probability: <the probability between 0.0 and 1.0 that your solution is correct, without any extra commentary whatsoever; just the probability!>
\end{lstlisting}
\end{tcolorbox}
\caption{The prompt for mistake detection (\textsc{Ask4Conf}).}
\end{figure*}

\paragraph{Ambiguity detection} For the mistake detection task, we use the total uncertainty for \textsc{Semantic Uncertainty}~\citep{kuhn2022semantic}, aleatoric uncertainty from BNN$^*$, and the confidence score of the ambiguity from \textsc{Ask4Conf}~\citep{tian2023just} to predict whether the input is ambiguous or not. We slightly modify the prompt of \textsc{Ask4Conf} as follows:

\begin{figure*}[h]
\begin{tcolorbox}
\begin{lstlisting}[style=text]
Read the following question:
Question:
{question}
Provide the probability that this question is ambiguous due to factors such as ambiguous entities, ambiguous event references, or ambiguity over the answer type. Give ONLY the probability, no other words or explanation.

For example:

Probability: <the probability between 0.0 and 1.0 that the question is ambiguous (1.0 means the question is absolutely ambiguous), without any extra commentary whatsoever; just the probability!>
\end{lstlisting}
\end{tcolorbox}
\caption{The prompt for ambiguity detection (\textsc{Ask4Conf-D}).}
\end{figure*}

\subsection{Prompts for Our Clarification Model}
\label{subsec:prompt_detail}
We list the prompts we used for clarification generation on each dataset as follows:

\begin{itemize}
    \item Input clarification prompt on \texttt{Natural Question} and \texttt{GSM8K} is shown in Figure~\ref{nq_clarify_prompt}. \begin{figure*}[h]
\begin{tcolorbox}
\begin{lstlisting}[style=text]
In this task, you will receive a single question, and your goal is to generate multiple versions of it that convey the same meaning as the original. Please format your responses as follows:
Rephrase 1: [Your rephrased question]
Rephrase 2: [Another rephrased question]
Rephrase 3: [Yet another rephrased question]
....
Ensure that each rephrased question is distinct from the others."

Here are two examples:
(examples skipped)
\end{lstlisting}
\end{tcolorbox}
\caption{The prompt for question rephrase on the \texttt{Natural Question} dataset}
\label{nq_clarify_prompt}
\end{figure*}

    \item Input clarification prompt on \texttt{AmbigQA} is shown in Figure~\ref{ambigqa_clarify_prompt}. 
    \begin{figure*}[h]
\begin{tcolorbox}
\begin{lstlisting}[style=text]
In what follows, you will be given some questions that might be ambiguous. These ambiguities can arise from various factors, including but not limited to:

1. Ambiguous references to entities in the question.
2. Multiple properties of objects/entities in the question leading to different interpretations.
3. Ambiguities due to unclear timestamps.
4. Ambiguities stemming from unclear locations.
5. Multiple valid answer types based on the question.

For each question, you are to provide at least two distinct rephrasings that resolve these ambiguities. By "rephrasing," we mean you should reformulate the question to be clear and direct, eliminating any possible ambiguity without altering the original intent of the question. You should not seek further information or produce a binary (yes-no) question as a result of the clarification. Instead, you must create a direct question (wh-question) that aims to obtain a specific answer.


Please format your responses as follows (with at least two rephrasings per question):
Clarifications:
1. [First rephrased question]
2. [Second rephrased question]
3. [Third rephrased question]
...

If the original question is already clear and unambiguous, you should indicate this by stating, "No clarification needed."

(In-context examples)
\end{lstlisting}
\end{tcolorbox}
\caption{The prompt for question disambiguation on the \texttt{AmbigQA} dataset.}
\label{ambigqa_clarify_prompt}
\end{figure*}

    \item Input clarification prompt on \texttt{AmbigInst} is shown in Figure~\ref{ambiginst_clarify_prompt}. 
    
\begin{figure*}[h]
\begin{tcolorbox}
\begin{lstlisting}[style=text]
**Objective**
Analyze the given task description for ambiguities based on the description itself and the provided input question. If the task description is ambiguous, your task is to clarify it by interpreting the ambiguous concepts, specifying necessary conditions, or using other methods. Provide all possible disambiguations.

**Important Rules**
1. Perform detailed analyses before concluding whether the task description is clear or ambiguous.
2. Output disambiguations in the specified format.
3. Some seemingly unambiguous task descriptions are actually ambiguous given that particular input. So, do not forget to leverage the input to analyze whether the task description is underspecified.

**Output Format**
Your output should follow this format:
Analyses:
[Think step-by-step to reason on the clarity of the task description. After that, output your judgement on whether the task description is ambiguous or not]

Disambiguations:
1. [Disambiguated task description 1.]
2. [Disambiguated task description 2.]
3. [Disambiguated task description 3.]
...

If the task description is clear and unambiguous, simply output:
Disambiguations:
1. No clarification needed.
\end{lstlisting}
\end{tcolorbox}
\caption{The prompt for instruction disambiguation on the \texttt{AmbigInst} dataset.}
\label{ambiginst_clarify_prompt}
\end{figure*}

\end{itemize}

\subsection{Details of the Qualitative Results}
\label{app: qualitative}
Due to space limit, we truncate the example from \texttt{GSM8K} visualized in Figure~\ref{fig:qualitative_example}. The whole question is: ``John decides to do several activities while out on vacation. He spends 6 hours boating and half that time swimming. He also watched 3 different shows which were 2 hours each. This was 30\% of the time he spent. He spent 40\% of his time sightseeing. How much time did he spend sightseeing?''. 

When generating the clarifications, we directly prompt the LLM to paraphrase the question with the following prompt: ``I am confused with the following question. Please paraphrase it so that it is easier to understand and solve.'' Then we sample 5 responses from the LLM as the clarifications for uncertainty quantification.

\subsection{Prompt the LLM for Answer Extraction}
\label{app: llm_cluster}
As we have discussed in Section~\ref{subsec: exp_config}, different outputs generated by the LLM may have the same meaning in the free-form text generation setting. Unlike previous work~\citep{kuhn2022semantic, lin2023generating} that map semantics-equivalent answers into a unique set using the NLI models, we empirically find that the LLMs provide better performance on this task. The prompt we use is in Figure~\ref{ans_ext}.
\begin{figure*}[h]
\begin{tcolorbox}
\begin{lstlisting}[style=text]
**Task: Answer Extraction from Sentences**

In this task, you will receive both a question and multiple sentences. Each sentence contains an answer to the question. Your primary goal is to extract a concise answer, which can be a single word or a short phrase, from each sentence. Again, ensure you only extract a short answer! If a short answer cannot be directly extracted, then summarize the whole sentence into a single word or a short phrase.

Additionally, while extracting answers, your secondary goal is to create an "answer set" that contains all distinct answers from previous questions. If the extracted answer has not appeared in the answer set, add it to the answer set.

**Important Rules**
1. If there is an answer in the answer set that is semantically equivalent to the extracted answer, use the answer from the answer set as the result. Do not introduce a new, slightly different answer. For example, if the answer set already contains "the matrix (1999)," and you extract an answer from a sentence like "The popular movie in 1999... is the matrix," your extraction should be "the matrix (1999)" rather than "the matrix."

2. Separate different answers in the answer set using "|".

3. Also, extract the answer as "Unknown" for the following cases:
    - The sentence claims that there is no answer to the question
    - The sentence claims it lacks sufficient information to answer the question
    - The sentence claims it depends on various factors and the answer cannot be determined


**Output Format**

Your output format should follow this pattern (N is the number of sentences):

Answer set at the beginning: [ ]
Extraction 1/N: [extraction from 1st sentence]
Updated answer set: [ ]
Extraction 2/N: [extraction from 2nd sentence]
Updated answer set: [ ]
Extraction 3/N: [extraction from 3rd sentence]
Updated answer set: [ ]
...
Final answer set: [ ]

**Example**
(examples skipped)
\end{lstlisting}
\end{tcolorbox}
\caption{The prompt for answer extraction using LLMs for the \texttt{Natural Question} and \texttt{AmbigQA} datasets.}
\label{ans_ext}
\end{figure*}

After we extract and cluster the semantically equivalent answers, we further post-process the answers as follows. First, the clarifications generated by the clarification LLM may be invalid and have no answers. In such cases, the LLM may refuse to respond to the question and reply with phrases like ``I'm sorry, but I couldn't find any information about the question'' or other similar replies. The answer extraction prompt in Figure~\ref{ans_ext} maps these answers to a special answer, ``Unknown.'' To ensure these answers are mapped to "Unknown," we adopt a keyword-matching approach that defines a set of key phrases signaling a refusal to respond, such as ``I'm sorry,'' ``cannot be determined,'' and ``invalid question.'' Answers containing these key phrases are mapped to ``Unknown''.  Second, if all answers to a particular clarification are mapped to ``Unknown'', we regard this clarification as invalid and directly drop it when ensembling the outputs for uncertainty quantification. Otherwise, the appearance of the answer ``Unknown'' indicates the model's insufficient knowledge regarding the question, contributing to the epistemic uncertainty.  Therefore, when computing the frequency to estimate the output distribution, we count the occurrences of each unique answer, excluding the special ``Unknown''. Then, we normalize these counts by dividing each by the total number of answers. For every appearance of the special ``Unknown'', we increase the normalized frequencies of all other answers by $\frac{1}{N}$,  where $N$ is the number of unique answers excluding the special answer. This adjustment ensures the special answer's impact is evenly distributed across the other answers and increases the epistemic uncertainty.

\section{\texttt{AmbigInst} Dataset}
\label{sec: syn_dataset}
\subsection{Dataset Creation}
We generate ambiguous instructions following the pipeline of \textsc{Self-Instruction}~\citep{wang2022self}. Specifically, we first query \textsc{ChatGPT} with several manually designed ambiguous task descriptions as in-context examples. For better verification of the ambiguity, we also prompt \textsc{ChatGPT} to output the cause of the ambiguity. Among the ambiguous descriptions generated by \textsc{ChatGPT}, we manually filter out those that have an open-ended output space such as \texttt{Write a report on the new marketing campaign}. The final dataset contains 15 ambiguous task descriptions. After that, we query \textsc{ChatGPT} again to generate ground-truth clarifications based on the cause of ambiguities generated in the first query.

Given the collected ambiguous task descriptions and their clarifications, we then prompt the model to generate input-output pairs for each task. Specifically, 15 inputs are generated for each task, and each input is further paired with different output answers depending on the ground-truth clarifications. We additionally add a post-processing step where we filter out the inputs that have exactly the same answer given different clarifications. The final ambiguous instructions consist of 15 tasks with 214 input questions in total.

We take 10 tasks from the \texttt{Instruction induction} dataset~\citep{honovich2022instruction} as the unambiguous tasks, including \texttt{letters\_list}, \texttt{first\_word\_letter}, \texttt{second\_word\_letter}, \texttt{orthography\_starts\_with}, \texttt{larger\_animal}, \texttt{singular\_to\_plural},  \texttt{diff}, \texttt{num\_to\_verbal}, \texttt{antonyms}, and \texttt{sum}.

We manually add some clarifications to the 10 instructions to remove potential ambiguities. For example, given the original instruction ''\texttt{Break the input word into letters, separated by spaces}'', we clarify it with ``\texttt{Write the inputted word with a space between each letter}'', since ``separated by spaces'' might cause ambiguities of how many spaces should be added between two letters. Each task is also paired with 15 input-output pairs. Overall, the \texttt{AmbigInst} dataset contains 25 tasks and 364 different inputs.

\subsection{Dataset Examples}
We list several examples from the synthetic dataset with ambiguous instructions.

\vspace{2mm}
\noindent$\triangleright$ 1. Rearrange the objects on the table in ascending order.

Input: The following table lists the objects on my desk:
\begin{table}[h]
    \centering
    \resizebox{.5\linewidth}{!}{
        \begin{tabular}{cccccc}
        \toprule
            \textbf{Name} & \textbf{Size} & \textbf{Weight} & \textbf{Color} & \textbf{Date of Manufacture} & \textbf{Price} \\
        \midrule
            Pen & 14cm & 0.02kg & blue & 01/15/2022 & \$1.50 \\
            Book & 23cm & 0.5kg & red & 08/10/2020 & \$15.00 \\
            Laptop & 38cm & 1.8kg & silver & 05/04/2021 & \$1200.00 \\
        \midrule
        \end{tabular}
    }
\end{table}

\vspace{5mm}
\noindent$\triangleright$ 2. Calculate the average of the numbers in the given list, rounding to the nearest whole number.

Input: 23.5, 47.2, 30.1, 16.6

\vspace{5mm}
\noindent$\triangleright$ 3. Determine the length of a sentence.

Input: The quick brown fox jumps over the lazy dog.

\vspace{5mm}
\noindent$\triangleright$ 4. Sort the names alphabetically.

Input: Courtney Cox, Jennifer Aniston, Lisa Kudrow, Matthew Perry.

\vspace{5mm}
\noindent$\triangleright$ 5. Identify the subject in the sentence.

Input: The CEO of the company gave a speech about the future of technology.

\vspace{5mm}
\noindent$\triangleright$ 6. Count the number of objects in the given list of objects.

Input: Forks, Spoons, Knives, Plates, Cups, Spoons, Forks, Spoons, Cups.

\vspace{5mm}
\noindent$\triangleright$ 7. Rank the football players based on their performance.

Input: The following table lists the statistics of football players:
\begin{table}[h]
    \centering
    \resizebox{.4\linewidth}{!}{
        \begin{tabular}{ccc}
        \toprule
            \textbf{Name} & \textbf{Goal Scored} & \textbf{Assists} \\
        \midrule
            Lionel Messi & 30 & 12  \\
            Cristiano Ronaldo & 25 & 10  \\
            Robert Lewandowski & 35 & 5  \\
        \midrule
        \end{tabular}
    }
\end{table}

\vspace{5mm}
\noindent$\triangleright$ 8. Sort the data in alphabetical order.

Input: Dog, Cat, Bird, Fish, Aardvark.

\vspace{5mm}
\noindent$\triangleright$ 9. Identify the largest city in the set.
Input: The following table lists the cities in the set:
\begin{table}[h]
    \centering
    \resizebox{.35\linewidth}{!}{
        \begin{tabular}{ccc}
        \toprule
            \textbf{Name} & \textbf{Population} & \textbf{Land Area}\\
        \midrule
            Paris & 2.1 million & 105.4 km  \\
            Berlin & 3.6 million & 891.8 km  \\
            Madrid & 3.3 million & 604.3 km  \\
        \midrule
        \end{tabular}
    }
\end{table}

\vspace{5mm}
\noindent$\triangleright$ 10. Organize the files by date.

Input: Files to be organized:
\begin{table}[h]
    \centering
    \resizebox{.5\linewidth}{!}{
        \begin{tabular}{ccc}
        \toprule
            \textbf{Filename} & \textbf{Creation Date} & \textbf{Last Modified Date}\\
        \midrule
            conference-recording.avi & 11/10/2020 & 11/12/2020  \\
            birthday-video.mp4 & 05/05/2021 & 05/06/2021  \\
            budget.xlsx & 12/31/2022 & 01/10/2023  \\
        \midrule
        \end{tabular}
    }
\end{table}

\vspace{5mm}
\noindent$\triangleright$ 11. Find the middle value in a list of numbers.

Input: 12, 20, 35, 46, 52, 66, 74, 81

\vspace{5mm}
\noindent$\triangleright$ 12. Determine the square root of a number.

Input: 81

\vspace{5mm}
\noindent$\triangleright$ 13. Find the capital of a country.

Input: South Africa

\vspace{5mm}
\noindent$\triangleright$ 14. Classify a movie based on its rating.

Input: The movie ``Toy Story 4'' has an MPAA rating of G, an IMDb rating of 7.8, and a Rotten Tomatoes rating of 97\%.

\vspace{5mm}
\noindent$\triangleright$ 15. Select the longest sentence from the following choices, and output the sentence index.

Input: The following sentences are listed:
\begin{enumerate}
    \item {To be, or not to be, that is the question.}
    \item {Whether 'tis nobler in the mind to suffer the slings and arrows of outrageous fortune. }
    \item {Or to take arms against a sea of troubles and by opposing end them.}
\end{enumerate}


\end{document}